\documentclass[11pt]{article}

\usepackage[preprint]{acl}

\usepackage{times}
\usepackage{latexsym}

\usepackage[T1]{fontenc}

\usepackage[utf8]{inputenc}

\usepackage{microtype}

\usepackage{inconsolata}

\usepackage{graphicx}
\usepackage{booktabs,tabularx}
\usepackage{multirow}
\usepackage{subcaption}

\title{Like a Therapist, But Not: Reddit Narratives of AI in Mental Health Contexts}

\author{Elham Aghakhani \\
  Drexel University \\
  \texttt{ea664@drexel.edu} \\\And
  Rezvaneh Rezapour \\
  Drexel University \\
  \texttt{sr3563@drexel.edu} \\}

\begin{document}
\maketitle
\begin{abstract}
Large language models (LLMs) are increasingly used for emotional support and mental health–related interactions outside clinical settings, yet little is known about how people evaluate and relate to these systems in everyday use. We analyze 5,126 Reddit posts from 47 mental health communities describing experiential or exploratory use of AI for emotional support or therapy. Grounded in the Technology Acceptance Model and therapeutic alliance theory, we develop a theory-informed annotation framework and apply a hybrid LLM–human pipeline to analyze evaluative language, adoption-related attitudes, and relational alignment at scale.
Our results show that engagement is shaped primarily by narrated outcomes, trust, and response quality, rather than emotional bond alone. Positive sentiment is most strongly associated with task and goal alignment, while companionship-oriented use more often involves misaligned alliances and reported risks such as dependence and symptom escalation. Overall, this work demonstrates how theory-grounded constructs can be operationalized in large-scale discourse analysis and highlights the importance of studying how users interpret language technologies in sensitive, real-world contexts.
\end{abstract}
\section{Introduction}\label{sec:introduction}

``Some absences keep their shape.'' When ChatGPT produced this reflection in response to a grieving psychologist, it startled him, not because of its eloquence, but for how closely it mirrored what he struggled to articulate himself. This encounter, described in the New York Times~\cite{lieberman2025therapist}, motivates a central question in this study: \textit{How do people evaluate, interpret, and engage with AI tools when they are used for emotional support or therapy?}, while illustrating the capacity of large language models (LLMs) to generate emotionally resonant language. 
Millions worldwide face barriers to traditional therapy, including high costs, limited availability of trained professionals~\cite{cnbc2021mentalhealth}, and the persistent stigma \cite{naslund2016future, thomson2024digital}. In response, systems such as ChatGPT, Claude, and Character AI are increasingly used as accessible, always-available conversational supports. This adoption is largely user-driven and occurs outside clinical settings, yet we lack systematic insight into how people evaluate, trust, and relate to these systems in real-world mental health contexts. 

Early work on conversational agents for mental health centered on scripted, rule-based systems such as ELIZA \cite{weizenbaum1966eliza} and PARRY \cite{colby1975artificial}, and later on CBT-oriented chatbots like Woebot and Wysa \cite{fitzpatrick2017delivering, beatty2022evaluating}. These systems delivered structured psychoeducation at scale but constrained engagement, whereas recent LLM-based systems enable open-ended, context-sensitive dialogue often perceived as empathetic or relational. However, most existing evaluations examine these systems in controlled settings, focusing on accuracy, safety, or therapeutic potential \cite{maurya2025assessing, thakkar2024artificial, roshanaei2025talk}, offering limited insight into how users interpret and negotiate AI's role in everyday mental health practices.

Online communities offer a valuable lens for examining these questions. Platforms such as Reddit host large-scale mental health discussions where users describe struggles, seek advice, and collectively interpret technologies, including the boundaries between AI support and human care~\cite{sit2024youth,bouzoubaa2023exploring}. 
Analyzing this discourse reveals how users assess the affordances and limitations of AI support and how these evaluations shape users' engagement with AI in relation to human mental health support.
To address these issues, we investigate the following research questions: \\
\textbf{RQ1:} How do individuals in online communities describe their use of AI for mental health support, including perceived functions, benefits, and risks?\\
\textbf{RQ2:} How are Technology Acceptance Model (TAM) dimensions associated with adoption-related attitudes toward AI for therapy and emotional support in online communities?\\
\textbf{RQ3:} How do users express therapeutic alliance with AI, and how do task, goal, and bond alignment relate to engagement outcomes?

We address these questions by analyzing 4.7 million posts across 47 condition-focused Reddit communities between November 2022 and August 2025. Using a multi-stage filtering pipeline, we identified posts describing experiential or exploratory use of AI as a therapeutic tool, resulting in a curated dataset of 5,126 posts focused on AI-supported mental health care. We developed a theory-grounded annotation schema integrating constructs from the Technology Acceptance Model~\cite{davis1989technology} and therapeutic alliance theory~\cite{bordin1979generalizability} to capture evaluative language, adoption-related attitudes, and relational alignment, including pragmatic evaluations (e.g., usefulness, trust, outcome quality) and task, goal, and bond dimensions shaping engagement and perceived risk, in user discourse.

Our results show that sustained engagement with AI for mental health support is shaped primarily by demonstrable outcomes, trust, and response quality, rather than by emotional bonding alone. Notably, therapeutic alliance is conditional: positive engagement is strongly associated with task and goal alignment, whereas emotional bond alone shows a weak relationship with positive sentiment and frequently co-occurs with dependency and reported harm.

Our work makes three contributions: (1) a novel dataset capturing how users evaluate and relate to AI systems for emotional support; (2) theory-grounded operationalization of Technology Acceptance Model and therapeutic alliance constructs for large-scale NLP analysis of adoption attitudes and relational alignment; and (3) quantitative and qualitative evidence linking system properties to adoption sentiment and therapeutic alliance, informing the design and evaluation of AI-supported mental health tools. We will release the dataset, annotation guidelines, and code.
\section{Related Work}
\label{sec:relatedwork}
\subsection{AI for Mental Health Support}
Advances in AI are transforming mental health care, with applications ranging from therapy analysis \cite{aghakhani-etal-2025-conversation} to early diagnostic tools and mood monitoring to conversational agents that deliver therapeutic techniques in real time~\cite{thakkar2024artificial, dehbozorgi2025application, graham2019artificial, cummins2020artificial, vaidyam2019chatbots}. Early systems such as ELIZA~\cite{weizenbaum1966eliza} and PARRY~\cite{colby1975artificial} demonstrated the potential of simulated dialogue, while later tools like Wysa and Woebot applied structured therapeutic approaches at scale~\cite{inkster2018empathy, beatty2022evaluating, gamble2020artificial}.  
More recently, LLMs have enabled open-domain conversation, context-sensitive responses, and language that users may interpret as empathetic or supportive~\cite{yu2024experimental, sorin2024large, ojeda2024relevance}. Recent studies show LLMs can deliver accurate and empathetic content~\cite{maurya2025assessing}, support positive mental health~\cite{thakkar2024artificial}, and even rival human-generated supportive messages~\cite{young2024role}. A recent survey synthesized the therapeutic potential of LLMs, their limitations, and ethical risks~\cite{na-etal-2025-survey}. 
Beyond mental health, prior NLP research has examined trust \cite{xie2024can,pessianzadeh2025generative}, alignment \cite{naseem-etal-2025-alignment}, and user perceptions of AI systems \cite{hojer-etal-2025-research,razi2025not}, typically through controlled experiments, surveys, or benchmark-driven evaluations. Prior work rarely examines mental health contexts or everyday evaluative and relational judgments, leaving users’ interpretations of LLM-based support underexplored. 

\subsection{Online Mental Health Discourse}
Online platforms have become key spaces for mental health discussion, enabling people to share experiences, seek advice, and access peer support \cite{thomson2024digital, naslund2020social}. Their accessibility, anonymity, and asynchronous nature make them especially valuable for those facing barriers to traditional care, such as cost, geography, or stigma \cite{naslund2016future, merchant2022opportunities, andalibi2017sensitive, bouzoubaa2024words}. Reddit, in particular, has emerged as a valuable source due to its topic-specific subreddits \cite{marshall2024understanding, bouzoubaa-etal-2024-decoding}, pseudonymity~\cite{naslund2020social}, and community norms that encourage candid disclosure~\cite{rayland2023social}. While this work demonstrates the richness of online mental health discourse, it has largely focused on condition classification \cite{dinu-moldovan-2021-automatic}, crisis detection \cite{d-lewis-etal-2025-tapping}, and symptom monitoring \cite{alhamed-etal-2024-monitoring}, rather than on how users perceive and discuss emerging tools such as AI in therapeutic contexts. 
We address this gap by examining how AI tools are evaluated and positioned in naturalistic mental health discourse. 

\section{Method}
\label{sec:method}
\subsection{Data}
Reddit has been widely used in mental health research due to its scale and topic-specific communities~\cite{sit2024youth, jiang-etal-2020-detection}. To analyze discourse around AI-based mental health tools, we curated condition-focused subreddits guided by DSM-5 diagnostic categories~\cite{diagnostic2013statistical} (e.g., depression, anxiety, bipolar disorder) to ensure broad and systematic coverage. Two authors independently reviewed DSM-5 categories and verified subreddit mappings, and we additionally included general mental health subreddits used for cross-diagnostic support and discussion (e.g., \texttt{r/mentalhealth}, \texttt{r/TalkTherapy}).
We selected 47 subreddits varying in size, ranging from fewer than 10,000 to over 1 million members (see Table~\ref{tab:dsm5_subreddits}), and collected Reddit submissions (excluding comments) using the ArcticShift API\footnote{\url{https://github.com/ArthurHeitmann/arctic_shift}}. We collected posts published between November 30, 2022 (ChatGPT's public release), and August 15, 2025, resulting in 4,703,056 submissions.
We applied preprocessing to improve data quality: concatenating titles and bodies, removing deleted or duplicate posts, and excluding posts with fewer than 10 tokens (the shortest 10\%). After preprocessing, the dataset comprised 3,530,486 posts. 

\subsection{Identifying AI-Relevant Posts}
To identify posts discussing AI tools (e.g., ChatGPT, Character AI) for therapeutic or emotional support, we used a scalable, multi-stage relevance filtering pipeline. To avoid applying LLM-based classification to the full corpus, we first narrowed the search space using keyword-based retrieval, followed by LLM-based validation on a reduced candidate set. In the first stage, we randomly sampled 120,000 posts and used GPT-4o mini~\cite{openai2024gpt4omini} to classify whether posts referenced AI use for emotional support, therapy, or mental health–related guidance. From posts labeled as relevant, we extracted AI-related keywords and phrases. 
This process resulted in around 800 unique AI-related terms (e.g., ``AI-Powered,'' ``Bing Chat,'' ``CHAI Bot''), which were manually reviewed and deduplicated to produce a refined list of 146 keywords. Applying this list to the full dataset resulted in a high-recall corpus of 572,734 posts.

In the second stage, we validated relevance before scaling LLM-based filtering. We randomly sampled 10,000 posts from the filtered corpus and used GPT-4o mini to label each as relevant or not. Two authors independently reviewed 200 posts (100 relevant, 100 not), resolving disagreements through discussion, and iteratively refined the prompt until agreement with human judgments was high (Fleiss' $\kappa = 0.90$). We then applied the finalized prompt to all 572,734 keyword-filtered posts, resulting in 6,206 posts describing AI use for therapeutic or emotional support.

To further characterize content, we manually reviewed a random sample of 200 posts, and identified four categories: \textit{experiential} (users describing personal AI use for therapy or emotional support), \textit{exploratory} (seeking advice or others' experiences), \textit{advertisement} (promotional or developer-posted content), and \textit{irrelevant} (posts that did not meaningfully discuss AI in a mental health context). Two authors independently annotated 50 posts per category using the same definitions provided to the LLM, achieving strong agreement with model classifications (Fleiss' $\kappa = 0.78$); disagreements were resolved through discussion. Since our analysis focuses on how individuals perceive and describe AI as a therapeutic tool, we retained only \textit{experiential} and \textit{exploratory} posts for further analysis, resulting in a final dataset of 5,126 posts.



\subsection{Annotation Framework}
To analyze how users evaluate AI as a therapeutic tool, we ground our annotation framework in two established theories from psychology and HCI to capture both pragmatic and relational aspects of AI-mediated mental health support: 

\begin{table*}[ht]
\centering
\resizebox{0.95\textwidth}{!}{%
\small
\begin{tabular}{@{}p{0.5cm}p{3.2cm}p{9cm}p{8cm}@{}}
\toprule
\textbf{} & \textbf{Dimension} & \textbf{Criteria} & \textbf{Measurement} \\
\midrule

\multirow{3}{*}{\rotatebox{90}{TAM}}
& Perceived Usefulness
& Does the user describe the AI as helpful or useful for emotional/mental-health tasks?
& Categorical: \texttt{[useful|not\_useful|not\_mentioned]} +Desc. \\
& Perceived Ease of Use
& Does the user describe the AI as easy, convenient, or accessible to use?
& Categorical: \texttt{[easy|difficult|not\_mentioned]} 
+Desc. \\
& Intention to Continue
& Does the user state an intention to keep using AI in the future?
& Categorical: \texttt{[yes|no|not\_mentioned]} \\
\midrule

& Perceived Trust
& Does the user indicate the AI is trustworthy, reliable, or safe in a mental-health context?
& Categorical: \texttt{[trustworthy|untrustworthy|not\_mentioned]} + Desc. \\
\multirow{5}{*}{\rotatebox{90}{TAM (ext.)}}& Output Quality
& Does the user judge the AI's responses as high vs. low quality (e.g., empathetic, thoughtful vs. vague, inaccurate)?
& Categorical: \texttt{[good|poor|not\_mentioned]} + Desc. \\
& Result Demonstrability
& Are tangible/behavioral outcomes mentioned (e.g., better sleep, reduced anxiety)?
& Categorical: \texttt{[pos.\_results|neg.\_results|not\_mentioned]} + Desc. \\
& Social Influence
& Does the user mention influence from peers/Reddit/others to use AI?
& Categorical: \texttt{[present|absent]} + Desc. \\
& Perceived Risks
& Mentions of limitations, harms, or risks in using AI for mental health (e.g., inaccuracy, emotional detachment).
& Categorical: \texttt{[mentioned|not\_mentioned]} + Desc. \\
\midrule

\multirow{3}{*}{\rotatebox{90}{\parbox{2cm}{\centering Therapeutic\\Alliance}}}
& Bond
& Does the user feel emotionally supported/understood/safe with the AI?
& Categorical: \texttt{[strong|weak|not\_mentioned]} + Desc. \\
& Task
& Is the AI helping with therapeutic actions (e.g., reflection, coping, journaling)?
& Categorical: \texttt{[aligned|misaligned|not\_mentioned]} + Desc. \\
& Goal
& Does the AI align with the user's mental-health goals (recovery, stability, growth)?
& Categorical: \texttt{[aligned|misaligned|not\_mentioned]} + Desc. \\
\midrule

\multirow{5}{*}{\rotatebox{90}{Other}}
& Usage Intent
& the primary function or reason a user engages with AI in a mental health related context.
&  Desc. \\
& Comparison to Therapy
& How the AI compares to traditional therapy.
& Categorical: \texttt{[better|worse|complementary|not\_mentioned]} \\
& AI Tool Mentioned
& The specific AI tool that is mentioned by the user.
& Desc. \\
& Sentiment toward AI 
& Overall Sentiment toward users' AI experience 
& Categorical: \texttt{[positive|negative|neutral]} + Desc.\\ 
& Mental Health Condition
& The specific mental health condition that is mentioned by the user.
& Desc. \\
\bottomrule
\end{tabular}}
\caption{Dimensions, associated frameworks, definitions, and measurement type (categorical or free-text (Desc).}
\label{tab:dimensions}
\vspace{-0.5cm}
\end{table*}

\noindent\textbf{Therapeutic Alliance Theory. }Therapeutic alliance refers to the collaborative and affective bond between therapist and client and is a key predictor of treatment outcomes~\cite{bordin1979generalizability}. It is typically described through three interdependent components: \texttt{bond} (trust, empathy, and emotional connection), \texttt{task} (agreement on therapeutic activities), and \texttt{goal} (shared commitment to objectives). Although originally developed for human psychotherapy, the framework has been applied to digital interventions~\cite{vowels2024ai, beatty2022evaluating}. We adapt this framework to analyze how users describe relational alignment with AI systems in mental health contexts. 

\noindent\textbf{Technology Acceptance Model. }TAM explains technology adoption through core beliefs about \texttt{usefulness} and \texttt{ease of use}, which influence attitudes, intentions, and actual use~\cite{davis1989technology, fishbein1979theory}. Extensions have added dimensions such as \texttt{Output Quality}, \texttt{result demonstrability}, and \texttt{social influence}~\cite{venkatesh2000theoretical}, while research in health informatics highlights the importance of \texttt{perceived trust} and \texttt{perceived risk} in sensitive domains~\cite{su2013extending, dhagarra2020impact}. Building on this work, we include core and extended TAM to capture user evaluations of AI tools in mental health contexts, where trust, risk, and outcomes are salient. 
We operationalized an annotation schema using these theories. 
From TAM and its extensions, we used \texttt{perceived usefulness}, \texttt{ease of use}, \texttt{output quality}, \texttt{result demonstrability}, \texttt{social influence}, \texttt{perceived trust}, \texttt{perceived risks}, and \texttt{intention to continue}. From therapeutic alliance, we included \texttt{task}, \texttt{bond}, and \texttt{goal}.
To capture context-specific aspects of AI–therapy discourse, we included \texttt{comparison to human therapy}, \texttt{AI tool mentioned}, \texttt{mental health condition}, \texttt{sentiment}, and \texttt{usage intent}. Table~\ref{tab:dimensions} shows all dimensions and coding criteria.

\subsection{Data Annotation}
\noindent\textbf{Human Annotation.} 
To assess annotation reliability, two authors independently annotated a random sample of 100 posts from the final dataset across all 13 dimensions (Table~\ref{tab:dimensions}) using shared guidelines.
Inter-annotator agreement measured by raw agreement, Cohen's Kappa, and Gwet's AC1 to account for label imbalance, ranged from $0.74$ to $0.95$ (raw), $0.37$ to $0.60$ (Cohen's Kappa), and $0.64$ to $0.94$ ( Gwet's AC1), indicating substantial to near-perfect agreement. Disagreements were resolved through discussion, and the resulting labeledset served as ground truth for subsequent model evaluation.

\noindent\textbf{LLM-Assisted Annotation and Evaluation.} 
We evaluated multiple (closed and open) LLMs selected from leaderboards~\footnote{\url{https://llm-stats.com/}} of high-performing systems available at the time of the study. These included \texttt{GPT-5.2}~\cite{openai_gpt52_2025}, \texttt{Gemini-3-Pro}~\cite{deepmind_gemini_pro_2025}, and \texttt{Claude-Opus-4.5}~\cite{anthropic_claude_opus45_2025}, which are proprietary models, as well as \texttt{Kimi-K2-Instruct}~\cite{moonshotai_kimik2instruct_2025} and \texttt{Qwen3-Next-80B-A3B-Instruct}~\cite{qwen3next80b_a3b_instruct_2025}, which are open-source models. 
All models were prompted using the same structured annotation procedure, including task instructions, construct definitions, and a standardized JSON output format. A sample schema was included to constrain outputs. We set the temperature to zero for all models; for GPT-5.2, we additionally fixed the reasoning setting to a medium level.
The full prompt is provided in the Appendix \ref{sec:annotation_prompt}.

For all dimensions, models produced free-text rationales for each label. We manually reviewed these outputs on the 100-post evaluation set to verify consistency with annotation definitions and post content.



\subsection{Thematic Analysis}
To facilitate a more fine-grained analysis of AI engagement and concerns (RQ1), we conducted additional thematic analysis on the LLM-generated free-text descriptions for two of our dimensions: \texttt{perceived risks} and \texttt{usage intent}. We applied LLM-guided thematic analysis~\cite{dai-etal-2023-llm} to consolidate these free-text descriptions into coherent, interpretable categories by grouping semantically similar phrases under shared themes. For example, risk descriptions such as ``compulsive use,'' ``overreliance,'' and ``obsessive use'' were consolidated into the category \textit{Addiction \& Dependence}. This process resulted in standardized taxonomies for both dimensions, enabling systematic analysis of engagement patterns and reported concerns across the corpus.

\section{Results}
\label{sec:results}
\subsection{Data Characteristics}

Our analysis includes 5,126 Reddit posts from 47 mental health communities (November 2022–August 2025) describing experiential (3,605) or exploratory (1,521) use of AI for therapeutic or emotional support. Engagement varied across communities, with anxiety-related subreddits showing the highest activity, i.e., nearly 500 posts in \texttt{r/Anxiety}. 
Trauma- and stressor-related communities were also prominent, led by \texttt{r/CPTSD} (422 posts). Substantial activity was observed in obsessive-compulsive (\texttt{r/OCD}, 392 posts) and neurodevelopmental communities (\texttt{r/ADHD} and \texttt{r/autism}, over 900 posts combined). General mental health subreddits (e.g., \texttt{r/therapy}, \texttt{r/therapists}, \texttt{r/mentalhealth}) also contained large volumes of AI-related discussions. Table~\ref{tab:data_characterictics} shows the distribution of posts across DSM-5 categories, and Figure~\ref{fig:temporal} shows their temporal trends. 

\subsection{Model Performance Across Dimensions}
We evaluated five LLMs on the human-annotated set across all categorical dimensions using precision, recall, and macro-averaged F1. Table~\ref{tab:model_scores_f1} reports F1 scores, with full precision and recall in Table~\ref{tab:model_scores}.
Across core and extended TAM dimensions, GPT-5.2 achieved the highest F1 scores on \texttt{perceived usefulness} (F1 = 0.72), \texttt{ease of use} (0.76), \texttt{perceived trust} (0.65), \texttt{output quality} (0.85), \texttt{result demonstrability} (0.82), \texttt{intention to continue} (0.72).
Gemini 3 Pro outperformed other models on \texttt{social influence} (0.82) and \texttt{perceived risks} (0.84). Performance on therapeutic alliance dimensions differed from other constructs and remained among the most challenging for all models. Gemini 3 Pro achieved the highest F1 scores for \texttt{bond} (0.78), \texttt{task} (0.71), and \texttt{goal} (0.64).

Based on these results, we adopted a hybrid annotation strategy for the full dataset of 5,126 posts: GPT-5.2 for TAM-based evaluative and outcome dimensions, and Gemini 3 Pro for therapeutic alliance, social influence, and perceived risk. Table~\ref{tab:construct_values} summarizes value distributions across dimensions and Table \ref{tab:example_posts_labels} shows examples of annotated data.
Most posts did not explicitly mention \texttt{bond}, \texttt{comparison to therapy}, \texttt{ease of use}, \texttt{perceived trust}, and \texttt{intention to continue} were frequently \texttt{not\_mentioned}. In contrast, \texttt{perceived usefulness}, \texttt{perceived risks}, \texttt{output quality}, and \texttt{result demonstrability} were more commonly expressed. \texttt{Sentiment} toward AI use was predominantly neutral, with positive sentiment more frequent than negative, while \texttt{social influence} was rarely mentioned. \texttt{Task} and \texttt{goal} alignment were more often aligned than misaligned when present.

\begin{table}[t]
\centering
\small
\setlength{\tabcolsep}{3pt}
\begin{tabular}{lccccc}
\toprule
\textbf{Dimension} &
\textbf{GPT} &
\textbf{Gem} &
\textbf{Cla} &
\textbf{Kimi} &
\textbf{Qwen} \\
\midrule
perceived\_usefulness   & \textbf{0.72} & 0.66 & 0.69 & 0.65 & 0.64 \\
ease\_of\_use           & \textbf{0.76} & 0.49 & 0.34 & 0.53 & 0.27 \\
perceived\_trust        & \textbf{0.65} & 0.64 & 0.64 & 0.49 & 0.51 \\
output\_quality         & \textbf{0.85} & 0.71 & 0.84 & 0.67 & 0.63 \\
result\_demonstrability & \textbf{0.82} & 0.67 & 0.79 & 0.69 & 0.59 \\
intention\_to\_continue & \textbf{0.72} & 0.66 & 0.70 & 0.59 & 0.41 \\
social\_influence       & 0.73 & \textbf{0.82} & 0.72 & 0.56 & 0.75 \\
perceived\_risks        & 0.70 & \textbf{0.84} & 0.80 & 0.77 & 0.70 \\
bond                    & 0.63 & \textbf{0.78} & 0.68 & 0.59 & 0.42 \\
task                    & 0.63 & \textbf{0.71} & 0.70 & 0.64 & 0.63 \\
goal                    & 0.57 & \textbf{0.64} & 0.64 & 0.40 & 0.56 \\
comparison\_to\_therapy & \textbf{0.64} & 0.63 & 0.59 & 0.61 & 0.41 \\
sentiment               & \textbf{0.77} & 0.68 & 0.75 & 0.70 & 0.62 \\
\midrule
\textbf{Macro F1}       & \textbf{0.70} & 0.68 & 0.67 & 0.60 & 0.59 \\
\bottomrule
\end{tabular}
\caption{F1 scores by dimension and model. Best score per dimension is shown in bold. GPT = GPT-5.2; Gem = Gemini~3~Pro; Cla = Claude~Opus~4.5.}
\label{tab:model_scores_f1}
\vspace{-0.5cm}
\end{table}

\begin{figure*}[t]
\centering
\label{fig:task}
\begin{subfigure}[b]{0.4\textwidth}
  \centering
  \includegraphics[width=\linewidth]{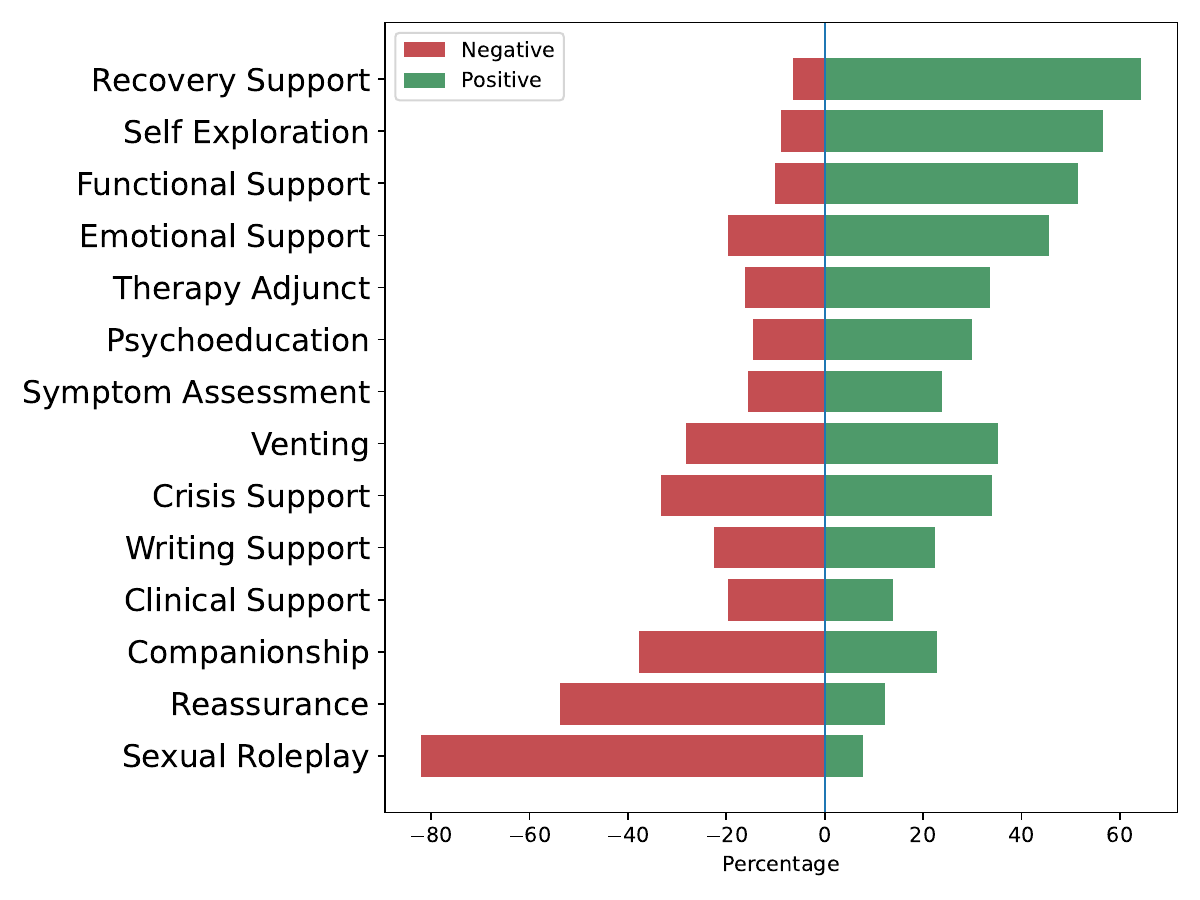}
  \caption{Sentiment distribution by task category}
  \label{fig:intent_sentiment}
\end{subfigure}
\hfill
\begin{subfigure}[b]{0.4\textwidth}
  \centering
  \includegraphics[width=\linewidth]{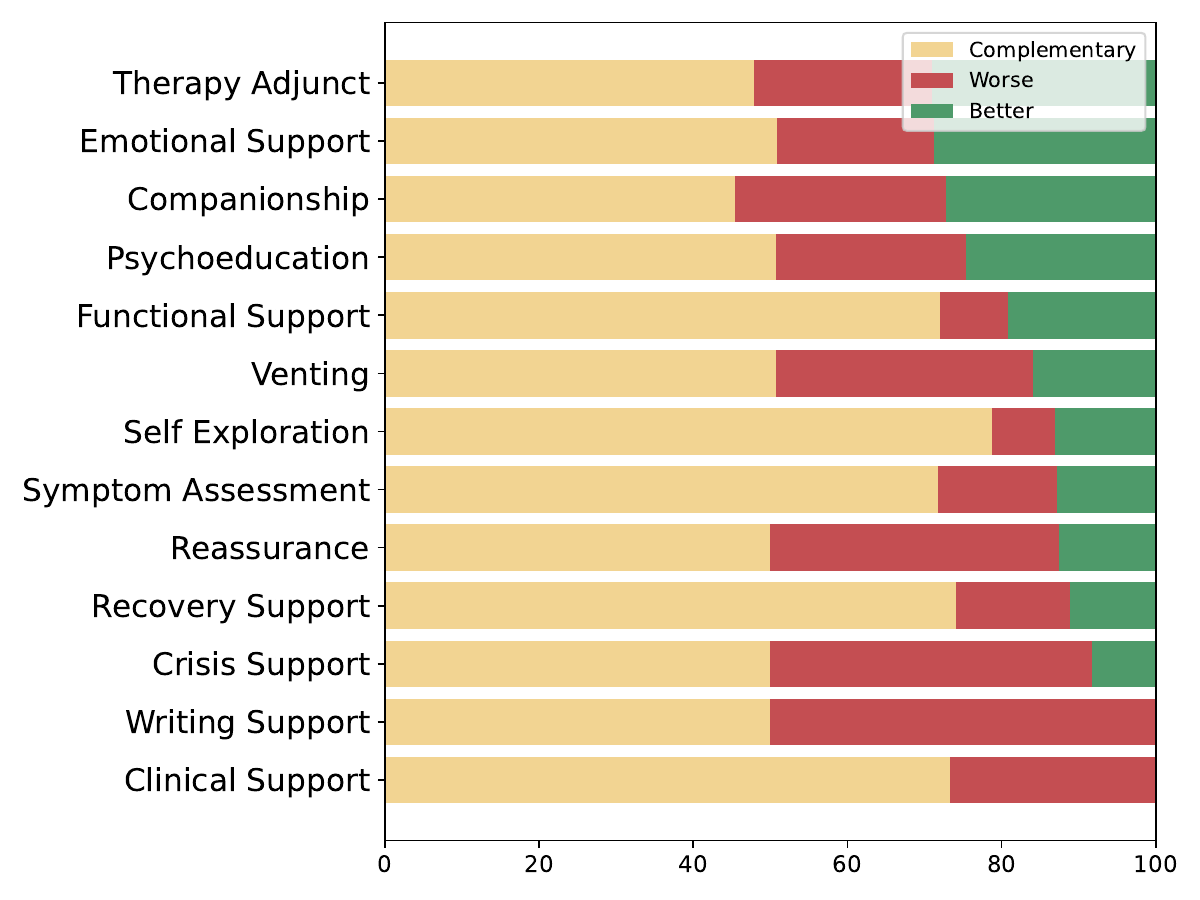}
  \caption{User evaluations of AI relative to human therapy by task category}
  \label{fig:comparison}
  \vspace{-0.25cm}
\end{subfigure}

\caption{Sentiment and comparative evaluations by AI usage category. Left: sentiment distribution by task. Right: AI positioning relative to human therapy.}
\label{fig:sentiment_comparison}
  \vspace{-0.5cm}
\end{figure*}
\subsection{RQ1: Patterns of AI Engagement in Mental Health Communities}
To characterize how AI is used and perceived in mental health contexts, we analyzed patterns of engagement reflected in users' discourse.

\noindent \textbf{Reported Usage Intent. }We examined reported \texttt{usage intent}s in relation to \texttt{sentiment} and \texttt{comparisons to human therapy}, and variation across \texttt{mental health condition}s. \texttt{usage intent} categories were derived via LLM-guided thematic analysis, which standardized free-text descriptions generated by LLM into a coherent taxonomy (e.g., Emotional Support, Venting, Companionship; see Table~\ref{tab:purposes} for the full list). 
Additional analysis of usage–condition co-occurrence is reported in the Appendix~\ref{sec:task_condition_heatmap}.

Across the corpus, \textit{Emotional Support} was the most prevalent (18.0\%) (Figure \ref{fig:intent_distribution}), encompassing empathy and emotional validation (Table \ref{tab:purposes}). \textit{Functional Support} (12.6\%) and \textit{Psychoeducation} (11.7\%) were also common, reflecting organizational assistance for ADHD and autism and learning about anxiety or coping strategies. Other common intents included \textit{Companionship} (9.4\%), \textit{Reassurance Seeking} (7.6\%), often linked to anxiety- or OCD-related validation, and \textit{Symptom Assessment} (7.3\%). \textit{Venting} (6.7\%) and \textit{Self Exploration} (6.5\%) appeared more often in CPTSD-related posts.

\noindent\textbf{Sentiment Across Tasks. }Sentiment toward AI varied across \texttt{usage intent}s. Intents involving sustained or reflective engagement, such as \textit{Recovery Support}, \textit{Self Exploration}, \textit{Functional Support}, and \textit{Emotional Support}, were more often associated with positive sentiment (users describing AI as helpful, validating, or confidence boosting), while \textit{Sexual Roleplay}, \textit{Reassurance Seeking}, and \textit{Companionship} showed higher negative sentiment, often reflecting guilt, symptom worsening, or emotional dependence. Task-level patterns are shown in Figures~\ref{fig:intent_sentiment} and~\ref{fig:comparison}.

\noindent\textbf{Comparison to Human Therapy. }Comparisons between AI and human therapy were relatively rare: 639 posts described AI as worse, 163 as better, and 478 as complementary. AI was most often framed as complementary, particularly for \textit{Functional Support}, \textit{Self Exploration}, and \textit{Recovery Support}. In contrast, \textit{Crisis Support} and \textit{Reassurance Seeking} were more frequently judged as worse, citing ineffective responses, safety concerns, or lack of appropriateness. Reports describing AI as better than therapy were uncommon and typically reflected barriers to accessing professional care.



\begin{figure}[t]
\centering
\includegraphics[width=0.9\columnwidth]{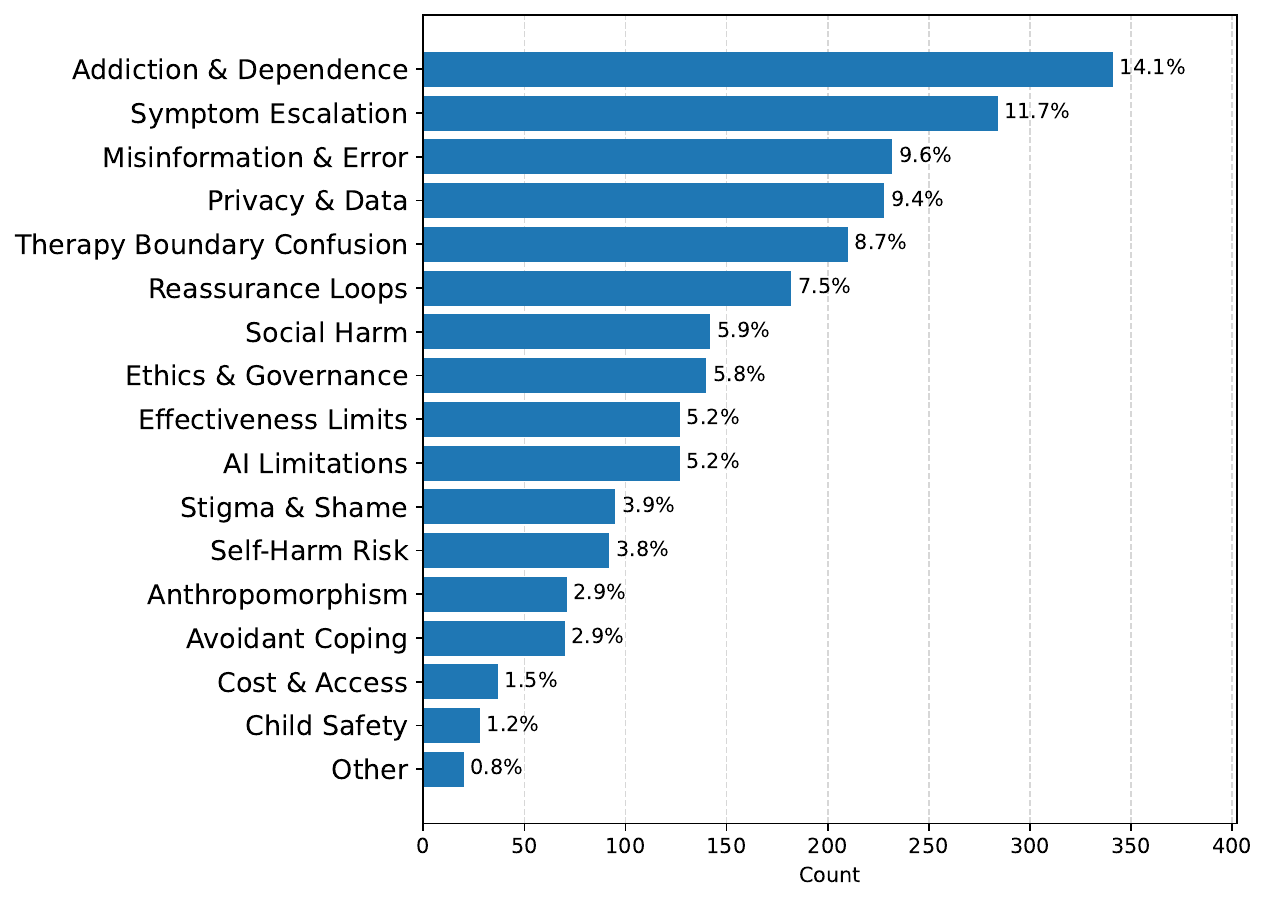}
\vspace{-0.25cm}
\caption{Reported Risks/Concerns}
\label{fig:risks}
  \vspace{-0.5cm}
\end{figure}

\noindent\textbf{Concerns/Risks. }
Alongside benefits, users frequently reported concerns about AI in therapeutic contexts: 2,637 of 5,126 posts (51\%) explicitly mentioned risks or limitations (Figure~\ref{fig:risks}). Using LLM-guided thematic analysis, we derived a standardized risk taxonomy (Table~\ref{tab:risks}).
The most common concern was \textit{Addiction and Dependence} (14.1\%), describing emotional reliance and compulsive use which co-occurred most strongly with \textit{Companionship} intent (Figure~\ref{fig:task_risk_heatmap}). \textit{Symptom Escalation} followed (11.7\%), including intensified anxiety, rumination, and trauma responses. 
\textit{Misinformation and Error} (9.6\%) was most strongly associated with \textit{Symptom Assessment} intent, reflecting concerns about incorrect interpretation or advice. \textit{Privacy and Data} risks (9.4\%) clustered around \textit{Clinical Support}, where users discussed documentation, records, or sensitive information handling. \textit{Reassurance Loops} were concentrated within \textit{Reassurance Seeking}, indicating a tight coupling between repeated reassurance and anxiety reinforcement. Although less frequent, \textit{Child Safety} risks appeared disproportionately in \textit{Sexual Roleplay}, and \textit{Stigma and Shame} and \textit{Self-Harm} appeared less often but raised concerns about judgment, suicidal ideation, or escalation of harm.


\subsection{RQ2: Adoption Pathways in AI for Mental Health}
Prior TAM work typically treats \texttt{intention to continue} as the outcome variable \cite{dhagarra2020impact, kim2012development}, but explicit intentions were sparse in our corpus: only 700 of 5,126 posts mentioned continued use, and 186 expressed no intention.
We therefore used sentiment toward AI as a proxy for adoption-related attitudes. Sentiment was more frequently expressed, with 2,198 neutral, 1,782 positive, and 1,120 negative. We treated positive sentiment as indicative of continued use and negative sentiment as indicative of discontinuation, excluding neutral cases.
We examined associations between TAM dimensions and sentiment using \texttt{chi-squared} tests and \texttt {Cram\'er's $V$}, limiting analyses to categorical dimensions and excluding \texttt{not\_mentioned} values per dimension.
As shown in Table~\ref{tab:tam_sentiment}, sentiment showed strong associations with
\texttt{result demonstrability}, \texttt{output quality}, and \texttt{perceived trust} (all $V > 0.90$, $p < .001$), and \texttt{perceived usefulness}
($V = 0.76$, $p < .001$). \texttt{Perceived risks} and \texttt{ease of use} showed moderate associations, while \texttt{social influence} was not significant.

\subsection{RQ3: Therapeutic Alliance and Its Role in AI Engagement}
We examined three therapeutic alliance dimensions in AI-mediated interactions: \texttt{task} alignment, \texttt{goal} alignment, and \texttt{bond}. We defined an \textit{overall therapeutic alliance} as
strongly aligned when \texttt{task} and \texttt{goal} were aligned and \texttt{bond} was strong, and as misaligned when \texttt{task} and \texttt{goal} were misaligned and \texttt{bond} was weak. Using this definition, 179 posts showed a strongly aligned alliance and 83 a misaligned one. Strongly aligned alliances were most common in \textit{Emotional Support} and \textit{Functional Support} tasks. In contrast, misaligned alliances appeared more frequently in companionship-oriented use. Posts describing a strong bond alongside negative sentiment frequently reported overattachment risks, including addiction and emotional dependence. 
Associations with sentiment were tested using \texttt{chi-squared} tests and \texttt {Cram\'er's $V$}. Strong associations were observed for \texttt{overall alliance} ($V = 0.95$), \texttt{task alignment} ($V = 0.93$), and \texttt{goal alignment} ($V = 0.79$), while \texttt{bond} showed a much weaker association ($V = 0.12$). Results are reported in Table~\ref{tab:alliance_sentiment}.


\begin{table}[t]
\centering
\resizebox{0.95\columnwidth}{!}{%
\begin{tabular}{p{0.5\columnwidth}cccc}
\hline
\textbf{Dimension} & \textbf{V} & \textbf{$\chi^2$} & \textbf{$p$} & \textbf{$n$} \\
\hline
Result demonstrability & 0.95 & 1906.51 & $< .001$ & 2,120 \\
Perceived trust        & 0.91 & 635.62  & $< .001$ & 764  \\
Output quality         & 0.90 & 1595.30 & $< .001$ & 1,981 \\
Perceived usefulness   & 0.76 & 1505.51 & $< .001$ & 2,612 \\
Perceived risks        & 0.43 & 543.38  & $< .001$ & 2,902 \\
Ease of use            & 0.40 & 86.62   & $< .001$ & 538  \\
Social influence       & 0.02 & 1.92    & .165     & 2,902 \\
\hline
\end{tabular}}
\caption{Associations between TAM dimensions and sentiment toward AI use for mental health support. Cramér’s V and chi-squared statistics are reported. Neutral sentiment cases were excluded.}
\label{tab:tam_sentiment}
  \vspace{-0.25cm}
\end{table}

\begin{table}[t]
\centering
\resizebox{0.95\columnwidth}{!}{%
\small
\setlength{\tabcolsep}{4pt}
\begin{tabular}{p{0.38\columnwidth}cccc}
\hline
\textbf{Dimension} & \textbf{V} & \textbf{$\chi^2$} & \textbf{$p$} & \textbf{$n$} \\
\hline
Overall alliance & 0.95 & 223.42  & $< .001$ & 249  \\
Task alignment   & 0.93 & 2101.80 & $< .001$ & 2414 \\
Goal alignment   & 0.79 & 1231.43 & $< .001$ & 1954 \\
Bond             & 0.12 & 12.66   & $< .001$ & 855  \\
\hline
\end{tabular}}
\caption{Associations between therapeutic alliance dimensions and sentiment toward AI use for mental health support, measured using chi-squared tests and Cram\'er's $V$. Neutral sentiment cases were excluded.}
\label{tab:alliance_sentiment}
  \vspace{-0.5cm}
\end{table}

\section{Discussion}
\label{sec:discussion}
\noindent \textbf{From Outcomes to Engagement.}
Across adoption-related analyses, result demonstrability emerged as the strongest predictor of adoption-related sentiment and continued use. Rather than abstract judgments of system quality, users expressed demonstrability through narrated change, accounts of improvement, deterioration, or ambivalence (e.g., better sleep, heightened rumination), a pattern well documented in online mental health communities \cite{andalibi2017sensitive, naslund2020social}. Although TAM extensions treat result demonstrability as a driver of adoption \cite{venkatesh2000theoretical, menzli2022investigation}, they typically operationalize it via surveys or task outcomes. Our results show that in naturalistic mental health discourse, adoption attitudes are embedded in narrative sensemaking grounded in \textit{felt changes} rather than abstract utility \cite{andalibi2017sensitive, de2013predicting}. These findings suggest prioritizing narrated outcomes in naturalistic discourse over preference or engagement metrics in NLP evaluation.

\noindent \textbf{When Emotional Bond Is Not Enough.}
We observe a clear asymmetry across therapeutic alliance dimensions: task and goal alignment are strongly associated with positive sentiment toward AI use, while emotional bond shows a much weaker relationship. Users describing AI as supporting structured, goal-directed psychological work reported more positive experiences, consistent with evidence that skills-oriented interactions better align with current AI capabilities~\cite{im2025clinical}.
In contrast, emotional bond was most common in companionship and reassurance-seeking contexts, where positive sentiment was less frequent and risks such as dependence, compulsive use, and symptom escalation were often reported \cite{babu2025digital}. These findings indicate that emotional bond alone is insufficient; when it forms without task and goal alignment, engagement may shift toward reassurance loops that sustain distress, as described in clinical models of anxiety \cite{starr2023dependency, rector2011assessing}. From an NLP perspective, this cautions against equating empathetic language with therapeutic suitability, as surface-level empathy can mask deeper misalignments and harms \cite{bender2021dangers, sharma2020computational, roshanaei2025talk}.

\noindent\textbf{Risk as an Emergent and Moral Property of AI Engagement.
}
Risks in AI-mediated mental health support are common and typically emerge from patterns of engagement rather than isolated failures. Over half of posts reference concerns, most often dependence, symptom escalation, misinformation, and privacy, described as cumulative trajectories involving repeated reassurance, emotional reliance, or difficulty disengaging. This aligns with prior human-centered NLP work showing that many harms arise from usage patterns rather than individual outputs \cite{blodgett2020language,ehsan2022human}.
Beyond psychological or informational harm, users' discourse reveals moral tensions around AI reliance. Many express guilt, shame, or self-judgment for turning to AI (e.g., ``pathetic,'' ``embarrassing,'' ``wrong''), even when reporting benefit, echoing prior work on stigmatized help-seeking and digital mental health use \cite{andalibi2017sensitive, naslund2016future}. 
Rather than dissatisfaction with responses, conflicted sentiment often reflects discomfort with reliance itself as users negotiate the legitimacy of AI support (``I know it's just an AI, but I have a safe place to talk''). Attending to such moralized expressions can surface early signs of potentially problematic engagement that may not be captured by sentiment or engagement metrics alone \cite{corrigan2014impact}.

\noindent \textbf{AI as Between-Session Mental Health Infrastructure.}
Users rarely frame AI as a replacement for professional therapy, instead describing it as complementary support when human care is unavailable, inaccessible, or insufficient—a pattern widely noted in digital mental health research \cite{bhatt2025digital}. This framing helps explain why functional support, self-exploration, and psychoeducation receive more positive evaluations, as these tasks align with AI's role as an auxiliary resource extending therapeutic work beyond the clinical encounter.
Conceptualizing AI as between-session mental health infrastructure clarifies both its value and its limits, shifting NLP research away from replacement narratives toward questions of reliability, boundary-setting, and support for self-directed work \cite{ehsan2022human}.
At the same time, the widespread use of general-purpose LLMs, particularly ChatGPT, as de facto mental health tools raises governance concerns: despite lacking health-specific safeguards, these systems are often treated as trustworthy, exposing users to privacy risks and blurred boundaries between informal support and professional care \cite{MITTechReview2025}. This gap between intended design and actual use underscores the need to study how people appropriate general-purpose NLP systems for mental health needs \cite{bender2021dangers, na-etal-2025-survey}.

Overall, our findings show that capturing key dynamics of AI-mediated mental health support is essential for understanding real-world human–AI engagement. Analyzing naturalistic user discourse with theory-informed constructs shows that in sensitive domains, how people \emph{live with} AI matters as much as what AI produces.

\section{Conclusion}
\label{sec:conclusion}
We studied how people evaluated and related to LLMs used for mental health support in naturalistic online discourse. Analyzing 5,126 Reddit posts across 47 mental health communities, we integrated Technology Acceptance Model and therapeutic alliance frameworks to examine adoption, evaluation, and relational alignment at scale. Engagement was driven primarily by perceived outcomes, trust, and response quality, with positive experiences most strongly associated with task and goal alignment rather than emotional bond alone. Users also reported concerns about dependence, symptom escalation, and misinformation, highlighting tensions between perceived support and potential harm.
\section{Limitations}
\label{sec:limitations}
This study relies on Reddit data, and findings should be interpreted in light of the platform's demographic and cultural biases. Subreddit-specific norms shape how mental health experiences and AI use are discussed, limiting generalizability to other populations, offline settings, or clinical contexts. Our analysis is also restricted to English-language posts, excluding perspectives from other linguistic and cultural settings. As with most analyses of online discourse, the data reflect self-selected users who choose to publicly discuss AI and mental health, and may overrepresent more salient, polarized, or reflective experiences.
We adapt constructs from the Technology Acceptance Model and therapeutic alliance theory, which were developed for survey-based studies and human psychotherapy. In this work, these frameworks serve as interpretive lenses for discourse analysis rather than formal tests of the theories themselves. Accordingly, our findings should not be read as validating or refuting these models, but as illustrating how their constructs surface in naturalistic user narratives.

Our annotation pipeline relies in part on large language models, introducing the possibility of misclassification, particularly for nuanced constructs such as sentiment and relational alignment. LLM-based annotation may also reflect normative assumptions that influence labeling. We mitigated these risks through human validation, agreement analysis, and conservative interpretation, but errors and biases may remain.
Finally, our analysis operates at the level of public discourse rather than longitudinal interaction traces or observed behavior. While Reddit posts capture rich evaluative and relational signals, they cannot directly reveal within-user trajectories, offline behavior, or causal effects of AI use over time. Importantly, this study does not evaluate mental health outcomes, therapeutic efficacy, or clinical safety. Self-reported experiences in public discourse should not be interpreted as evidence of benefit or harm at the individual or population level. Future work combining discourse analysis with longitudinal interaction data, interviews, or diary-based methods could more precisely characterize escalation, disengagement, and recovery dynamics. 

\section{Ethics Statement}
This study analyzes publicly available Reddit posts in which users discuss sensitive mental health experiences. We designed our methodology to minimize potential harm and respect user privacy. All data were collected from public subreddits in accordance with Reddit's terms of service, and we made no attempts to identify, profile, contact, or interact with individual users, including potentially vulnerable populations, nor to intervene in or influence ongoing discussions.
To reduce the risk of re-identification, we do not release raw post text. Instead, we will share post identifiers (e.g., Reddit IDs) and derived annotation labels, enabling reproducibility for researchers with appropriate data access while limiting exposure of sensitive content. All analyses were conducted at the aggregate level, and examples were paraphrased or abstracted to avoid revealing identifiable or distressing details.
We recognize the risk that findings from this work could be misinterpreted or misused to overstate the therapeutic value of AI systems or to justify their deployment as substitutes for professional mental health care. Our results are intended to inform responsible evaluation, design, and governance of AI systems in mental health contexts, not to recommend clinical use or automated intervention. The AI systems discussed in this study are not substitutes for trained clinicians and should only be used with appropriate safeguards, transparency, and ethical oversight.

\section{Acknowledgement}
\label{sec:ack}
We thank Darshit Rai for his support in data collection and for validating the relevancy annotations. His careful review and feedback played an important role in maintaining the quality of the annotated data. We also thank OpenAI for the research credits.  
\bibliography{aaai25}

\appendix

\section{Appendix}
\setcounter{table}{0}
\setcounter{figure}{0}

\renewcommand{\thetable}{A.\arabic{table}}
\renewcommand{\thefigure}{A.\arabic{figure}}

\subsection{Annotation Prompt}\label{sec:annotation_prompt}
\begin{figure*}[th]
\centering
\includegraphics[width=\textwidth]{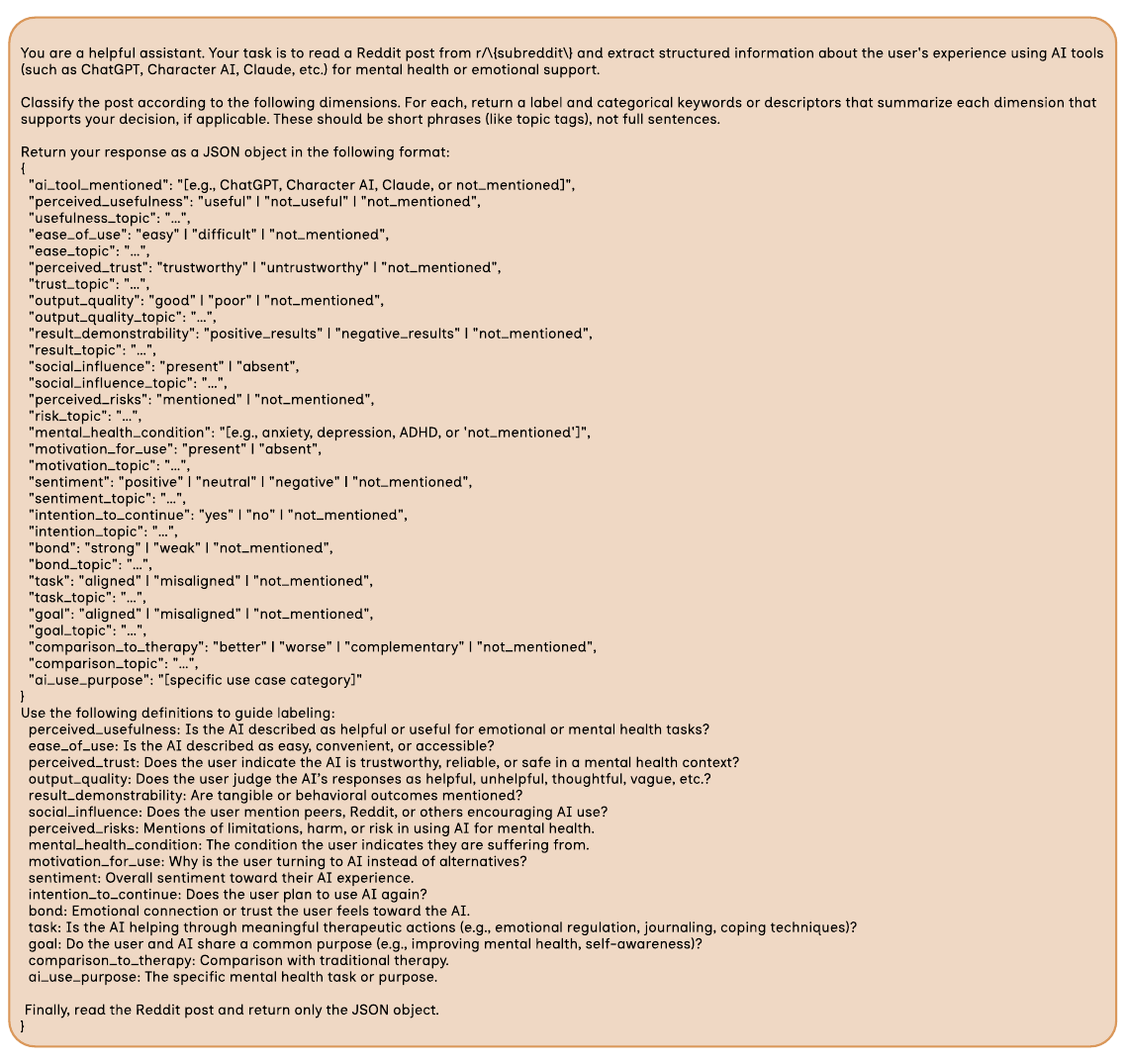}
\caption{The prompt used to extract dimensions from Reddit posts}
\label{fig:annotation_prompt}
\end{figure*}

Figure \ref{fig:annotation_prompt} presents the annotation prompt used to extract dimensions from Reddit posts.

\subsection{LLM Performance Across Annotation Dimensions}
\label{sec:llm_performance_full}

Table~\ref{tab:model_scores} reports precision, recall, and F1 scores for all
five evaluated LLMs across the full set of categorical annotation dimensions.
Models were evaluated against the human-validated reference set, and performance
varied substantially by dimension.

\begin{table*}[t]
\centering
\resizebox{0.95\textwidth}{!}{%
\small
\setlength{\tabcolsep}{4pt}
\begin{tabular}{lccc ccc ccc ccc ccc}
\toprule
\textbf{Dimensions} &
\multicolumn{3}{c}{\textbf{GPT 5.2}} &
\multicolumn{3}{c}{\textbf{Gemini 3 pro}} &
\multicolumn{3}{c}{\textbf{Claude-opus-4-5}} &
\multicolumn{3}{c}{\textbf{Kimi-K2-Instruct}} &
\multicolumn{3}{c}{\textbf{Qwen3}} \\
\cmidrule(lr){2-4}\cmidrule(lr){5-7}\cmidrule(lr){8-10}\cmidrule(lr){11-13}\cmidrule(lr){14-16}
& \textbf{P} & \textbf{R} & \textbf{F1}
& \textbf{P} & \textbf{R} & \textbf{F1}
& \textbf{P} & \textbf{R} & \textbf{F1}
& \textbf{P} & \textbf{R} & \textbf{F1}
& \textbf{P} & \textbf{R} & \textbf{F1} \\
\midrule
perceived\_usefulness      & 0.82 & 0.68 & \textbf{0.72} & 0.82 & 0.70 & 0.66 & 0.79 & 0.70 & 0.69 & 0.75 & 0.70 & 0.65 & 0.80 & 0.61 & 0.64 \\
ease\_of\_use              & 0.73 & 0.81 & \textbf{0.76} & 0.47 & 0.60 & 0.49 & 0.40 & 0.50 & 0.34 & 0.59 & 0.74 & 0.53 & 0.56 & 0.59 & 0.27 \\
perceived\_trust           & 0.73 & 0.67 & \textbf{0.65} & 0.63 & 0.70 & 0.64 & 0.64 & 0.67 & 0.64 & 0.50 & 0.59 & 0.49 & 0.61 & 0.73 & 0.51 \\
output\_quality            & 0.85 & 0.86 & \textbf{0.85} & 0.74 & 0.75 & 0.71 & 0.82 & 0.86 & 0.84 & 0.74 & 0.77 & 0.67 & 0.70 & 0.72 & 0.63 \\
result\_demonstrability    & 0.81 & 0.83 & \textbf{0.82} & 0.82 & 0.68 & 0.67 & 0.80 & 0.80 & 0.79 & 0.78 & 0.70 & 0.69 & 0.69 & 0.61 & 0.59 \\
intention\_to\_continue    & 0.85 & 0.66 & \textbf{0.72} & 0.62 & 0.75 & 0.66 & 0.67 & 0.79 & 0.70 & 0.58 & 0.74 & 0.59 & 0.51 & 0.65 & 0.41 \\
social\_influence          & 0.68 & 0.84 & 0.73 & 0.75 & 0.97 & \textbf{0.82} & 0.66 & 0.95 & 0.72 & 0.55 & 0.59 & 0.56 & 0.70 & 0.85 & 0.75 \\
perceived\_risks           & 0.75 & 0.78 & 0.70 & 0.84 & 0.88 & \textbf{0.84} & 0.80 & 0.85 & 0.80 & 0.79 & 0.84 & 0.77 & 0.74 & 0.77 & 0.70 \\
bond                       & 0.67 & 0.74 & 0.63 & 0.73 & 0.85 & \textbf{0.78} & 0.65 & 0.72 & 0.68 & 0.63 & 0.75 & 0.59 & 0.55 & 0.67 & 0.42 \\
task                       & 0.61 & 0.67 & 0.63 & 0.80 & 0.72 & \textbf{0.71} & 0.66 & 0.80 & 0.70 & 0.66 & 0.68 & 0.64 & 0.75 & 0.65 & 0.63 \\
goal                       & 0.60 & 0.58 & 0.57 & 0.70 & 0.66 & \textbf{0.64} & 0.71 & 0.65 & 0.64 & 0.40 & 0.43 & 0.40 & 0.69 & 0.60 & 0.56 \\
comparison\_to\_therapy    & 0.66 & 0.63 & \textbf{0.64} & 0.60 & 0.66 & 0.63 & 0.60 & 0.62 & 0.59 & 0.61 & 0.60 & 0.61 & 0.49 & 0.49 & 0.41 \\
sentiment                  & 0.78 & 0.77 & \textbf{0.77} & 0.78 & 0.69 & 0.68 & 0.78 & 0.74 & 0.75 & 0.81 & 0.73 & 0.70 & 0.68 & 0.66 & 0.62 \\
\midrule
\textbf{overall macro F1}  & \multicolumn{3}{c}{\textbf{0.70}} & \multicolumn{3}{c}{0.68} & \multicolumn{3}{c}{0.67} & \multicolumn{3}{c}{0.60} & \multicolumn{3}{c}{0.59} \\
\bottomrule
\end{tabular}}
\caption{Precision (P), recall (R), and F1 scores across models for each dimension.}
\label{tab:model_scores}
\vspace{-0.5cm}
\end{table*}
\subsection{Dimension Distributions Across the Corpus}
\label{sec:dimension_distribution}
After applying the hybrid annotation pipeline to the full set of 5,126 posts, categorical labels for all Technology Acceptance Model and therapeutic alliance dimensions were produced. Table \ref{tab:construct_values} presents the proportional distribution of categorical values for each dimension across the corpus. Across dimensions, not\_mentioned was common. For example, \texttt{intention to continue} was absent in 82.6\% of posts. When expressed, 13.7\% indicated continued or planned AI use, while 3.6\% reported discontinuation or intent to stop. Some dimensions were more frequently expressed. \texttt{Perceived usefulness} was coded as useful in 59.3\% of posts and not useful in 11.5\%. For \texttt{output quality}, 35.9\% described good quality and 11.0\% poor quality, while 53.1\% did not include an explicit judgment.

\begin{figure}[]
\centering
\includegraphics[width=\columnwidth]{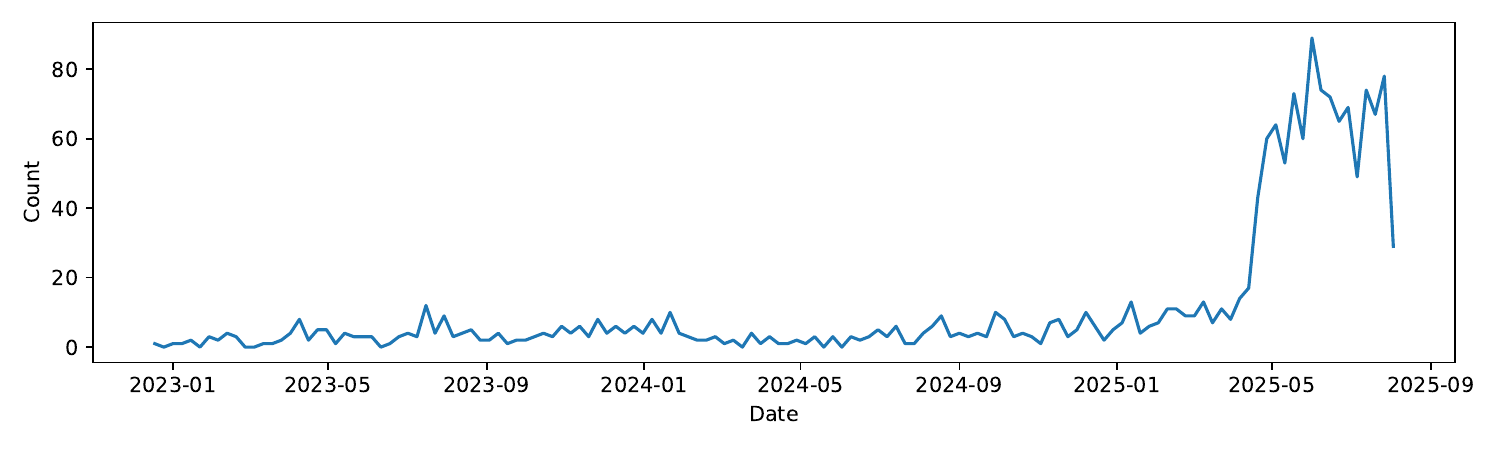}
\caption{Temporal distribution of AI-related mental health posts in the dataset.}
\label{fig:temporal}
\end{figure}


\subsection{Distribution of Usage Intent Categories}
\label{sec:usage_intent_distribution}

\begin{figure}[]
\centering
\includegraphics[width=\columnwidth]{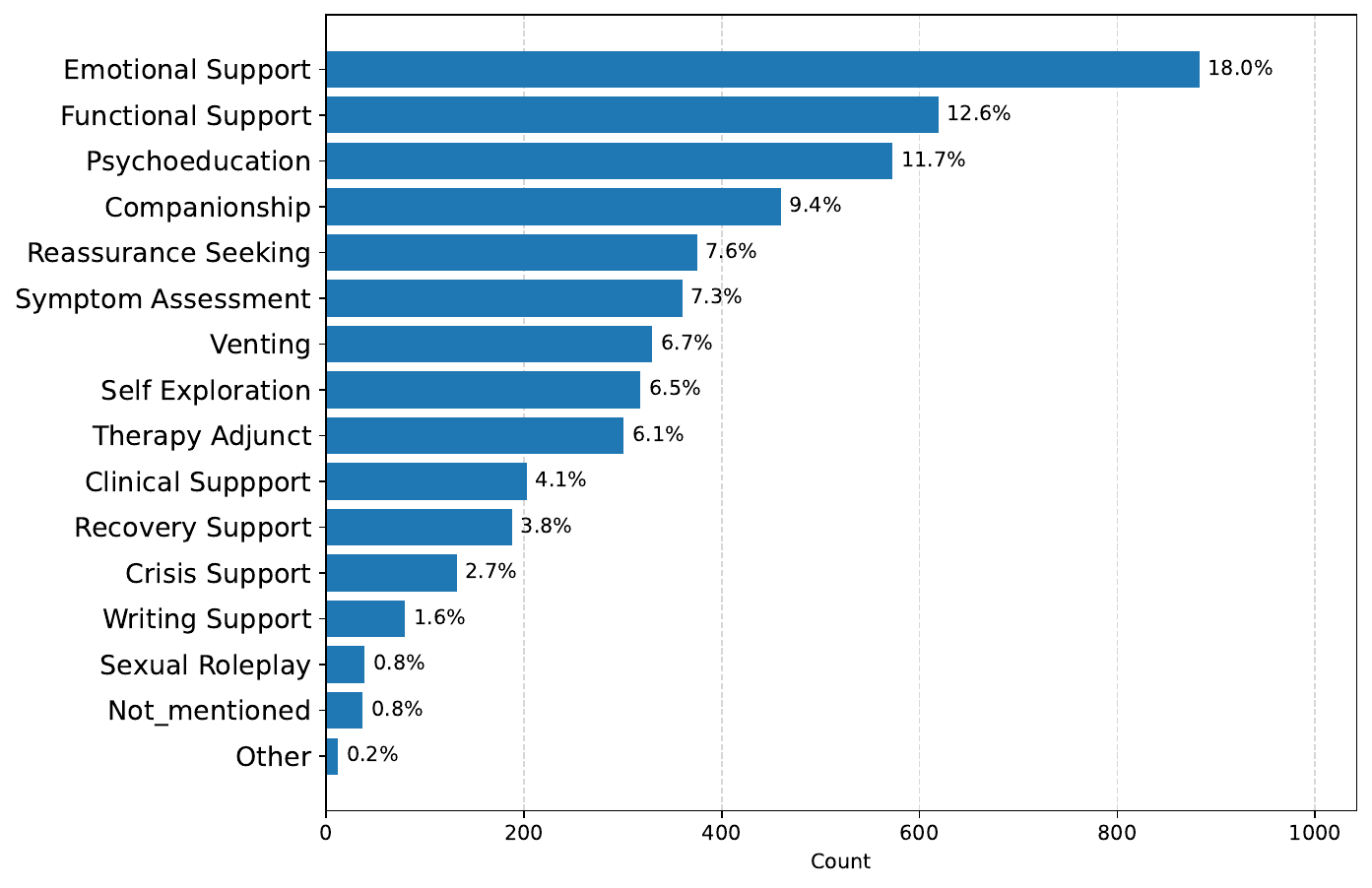}
\caption{Distribution of Usage Intent Categories Reported by Users}
\label{fig:intent_distribution}
\end{figure}

Figure \ref{fig:intent_distribution} shows the distribution of reported AI usage intents. \textit{Emotional
Support} is the most common use, followed by \textit{Functional Support} and
\textit{Psychoeducation}, indicating that AI is primarily used for ongoing
emotional validation and practical coping support. Relational intents such as
\textit{Companionship} and \textit{Reassurance Seeking} appear less frequently, while task-specific uses including \textit{Symptom Assessment}, \textit{Therapy Adjunct}, and \textit{Clinical Support} are relatively rare.

\subsection{Task–Condition Co-occurrence Pattern}
\label{sec:task_condition_heatmap}

\begin{figure*}[]
\centering
\includegraphics[width=\textwidth]{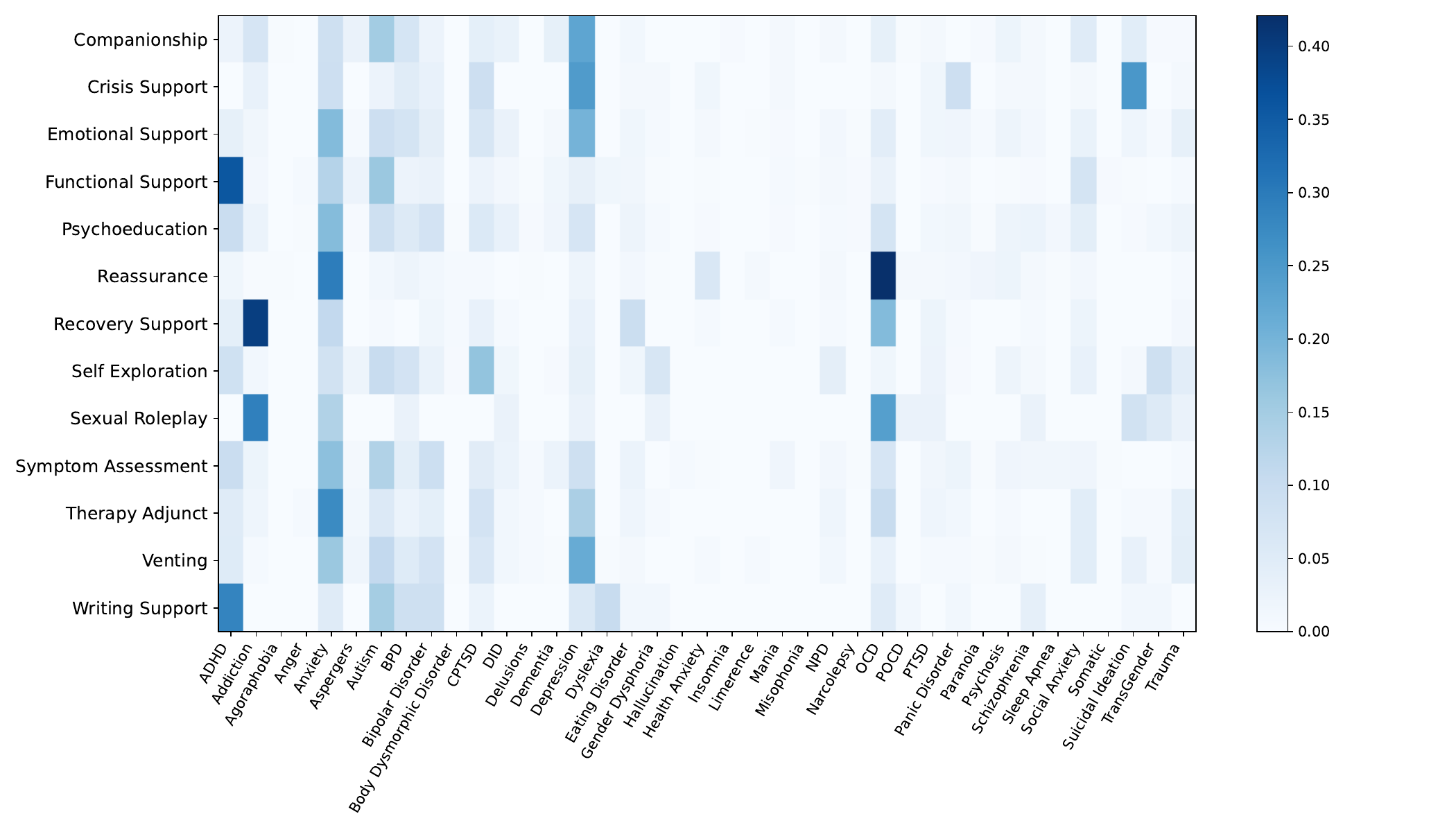}
\caption{Co-occurrence of Mental Health Conditions and Usage Intents.}
\label{fig:task_condition_heatmap}
\end{figure*}

Figure~\ref{fig:task_condition_heatmap} shows a heatmap of the co-occurrence between AI usage tasks and reported mental health conditions in the dataset. Rows correspond to usage task categories, and columns correspond to mental
health conditions. Mental health conditions were identified in two ways. When explicitly stated,
conditions were extracted from post text using LLM-based annotation. For posts in which no condition was explicitly mentioned, we used the associated subreddit
as a proxy for the relevant condition when applicable (e.g., posts from \texttt{r/Anxiety} mapped to anxiety-related conditions). This approach allowed
us to capture both self-reported and contextually inferred conditions while maintaining broad coverage across the corpus. Cell intensities reflect the relative frequency of each task–condition pairing, showing how different AI usage tasks are distributed across mental health contexts.

\subsection{Risk–Intent Co-occurrence Patterns}
\label{sec:task_condition_heatmap}
\begin{figure*}[]
\centering
\includegraphics[width=\textwidth]{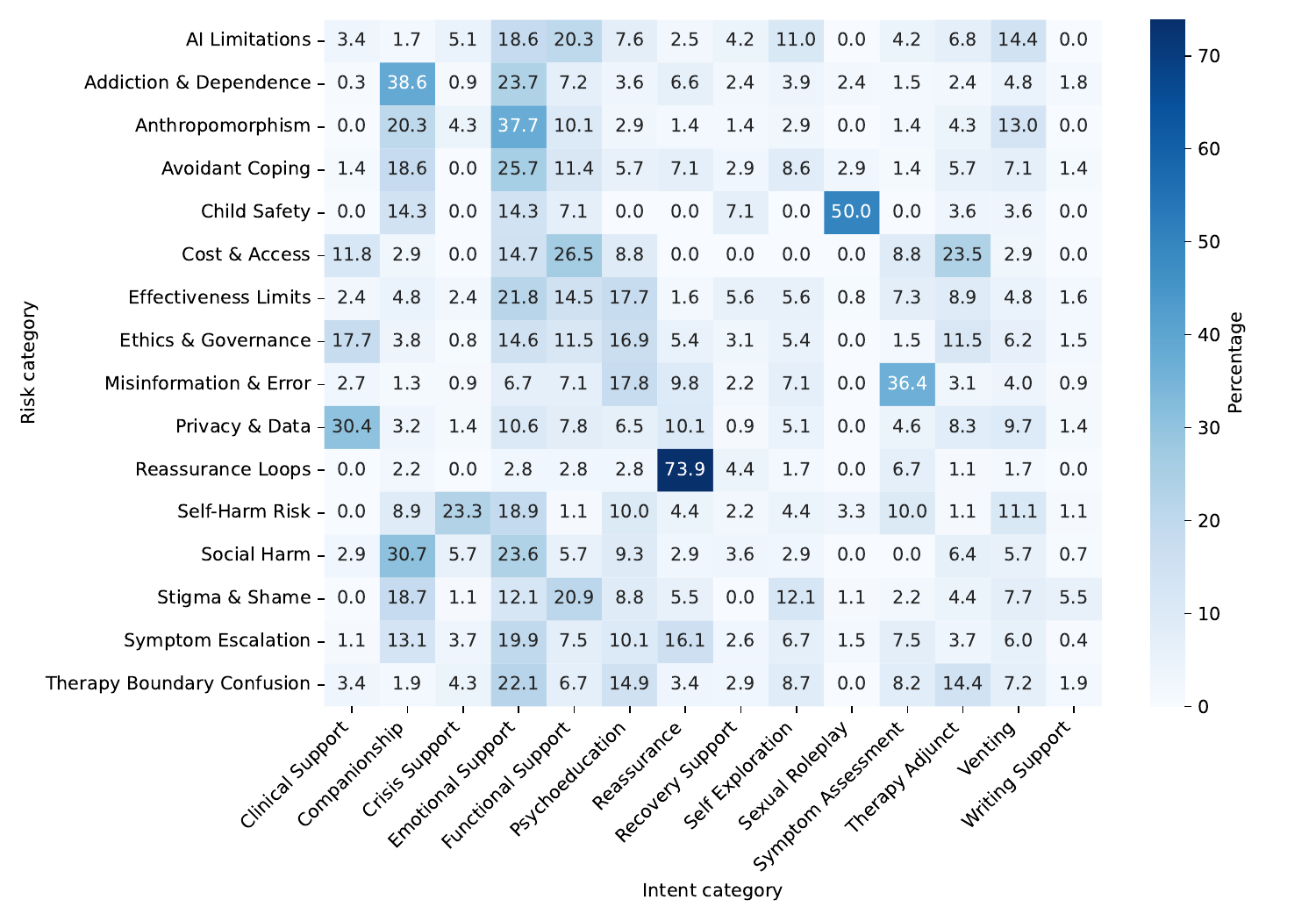}
\caption{Co-occurrence of Intent Categories within Risk Categories.}
\label{fig:task_risk_heatmap}
\end{figure*}

Figure~\ref{fig:task_risk_heatmap} shows the co-occurrence between AI usage intents
and reported risk categories. Percentages indicate the distribution of risk
mentions within each intent.
Risks cluster strongly by intent. \textit{Companionship} use is most often
associated with \textit{Addiction \& Dependence}, reflecting concerns about
emotional reliance. \textit{Reassurance Seeking} is dominated by
\textit{Reassurance Loops}. \textit{Privacy \& Data} concerns appear most
frequently in \textit{Clinical Support}, while \textit{Child Safety} risks are
concentrated in \textit{Sexual Roleplay}. \textit{Misinformation \& Error} is most
prominent in \textit{Symptom Assessment}.
Overall, the heatmap shows that risks are task-specific rather than uniform,
highlighting the importance of intent-aware safety evaluation in AI-mediated
mental health support.

\subsection{Categorization of Subreddits under DSM-5 Categories}
Table~\ref{tab:dsm5_subreddits} organizes the selected mental health subreddits into DSM-5 diagnostic categories. This categorization provides the clinical framing for our dataset and ensures coverage across a wide range of mental health conditions and community types.

\newcolumntype{Y}{>{\raggedright\arraybackslash}X}
\begin{table*}[t]
\centering
\small
\setlength{\tabcolsep}{4pt}
\begin{tabularx}{\textwidth}{YYY}
\toprule
\textbf{DSM-5 Category} & \textbf{Definition} & \textbf{Relevant Subreddits} \\
\midrule
Neurodevelopmental Disorders & Early developmental onset; impairments in personal, social, academic, or occupational functioning. & r/ADHD, r/autism, r/aspergers, r/Dyslexia \\
\midrule
Schizophrenia Spectrum and Other Psychotic Disorders & Disorders with psychosis, delusions, hallucinations, or disorganized thinking. & r/schizophrenia, r/Psychosis \\
\midrule
Bipolar and Related Disorders & Mood disorders with episodes of mania/hypomania and depression. & r/bipolar, r/bipolar2, r/BipolarReddit \\
\midrule
Depressive Disorders & Persistent sadness, emptiness, or irritability that significantly impairs functioning. & r/depression, r/depression\_help \\
\midrule
Anxiety Disorders & Excessive fear, worry, and related behavioral disturbances. & r/Anxiety, r/Anxietyhelp, r/socialanxiety, r/PanicAttack, r/HealthAnxiety \\
\midrule
Obsessive-Compulsive and Related Disorders & Obsessions, compulsions, or repetitive behaviors. & r/OCD \\
\midrule
Trauma- and Stressor-Related Disorders & Disorders following exposure to trauma or stress. & r/CPTSD, r/ptsd, r/trauma, r/SomaticExperiencing \\
\midrule
Dissociative Disorders & Disruptions in consciousness, memory, identity, or perception. & r/DID, r/Dissociation \\
\midrule
Feeding and Eating Disorders & Disturbances in eating behaviors that impair health or psychosocial functioning. & r/AnorexiaNervosa, r/EatingDisorders, r/BingeEatingDisorder \\
\midrule
Substance-Related and Addictive Disorders & Disorders related to the use of substances or addictive behaviors. & r/stopdrinking, r/Drugs, r/addiction \\
\midrule
Neurocognitive Disorders & Primary deficit is cognitive decline (memory, attention, language). & r/dementia, r/Alzheimers \\
\midrule
Sleep-Wake Disorders & Disorders affecting quality, timing, or amount of sleep. & r/SleepApnea, r/Narcolepsy \\
\midrule
Personality Disorders & Enduring patterns of inner experience and behavior that deviate from cultural expectations. & r/NPD, r/personalitydisorders, r/BPD \\
\midrule
Disruptive, Impulse-Control, and Conduct Disorders & Problems with emotional or behavioral self-control. & r/Anger \\
\midrule
Other / Transdiagnostic or General Mental Health & Not tied to a specific DSM-5 disorder but broadly related to mental health. & r/mentalhealth, r/MentalHealthSupport, r/therapists, r/therapy, r/TalkTherapy, r/selfimprovement, r/Mindfulness, r/Antipsychiatry, r/asktransgender, r/SuicideWatch \\
\bottomrule
\end{tabularx}
\caption{Categorization of selected mental health subreddits under DSM-5 categories.}
\label{tab:dsm5_subreddits}
\end{table*}

\subsection{Subreddit Dataset Statistics}
Table~\ref{tab:data_characterictics} summarizes the 47 selected subreddits organized by DSM-5 categories. For each subreddit, we report subscriber counts, the number of posts processed, and the number of posts deemed relevant to AI in mental health contexts.

\begin{table*}[t]
\centering
\small
\renewcommand{\arraystretch}{.99}
\begin{tabularx}{\textwidth}{YYYYY}
\hline
\textbf{DSM-5 Category} & \textbf{Subreddit} & \textbf{\#Subscribers} & \textbf{\#Processed Posts} & \textbf{\#AI Relevant Posts} \\
\hline
Neurodevelopmental Disorders & r/ADHD & 2.1m & 223{,}457 & 50 \\
 & r/autism & 477k & 223{,}075 & 158 \\
 & r/aspergers & 174k & 43{,}153 & 138 \\
 & r/Dyslexia & 34k & 5{,}692 & 21 \\
\hline
Schizophrenia Spectrum and Other Psychotic Disorders & r/schizophrenia & 94k & 57{,}330 & 87 \\
 & r/Psychosis & 66k & 26{,}417 & 60 \\
\hline
Bipolar and Related Disorders & r/bipolar & 262k & 82{,}815 & 38 \\
 & r/bipolar2 & 81k & 45{,}712 & 55 \\
 & r/BipolarReddit & 98k & 35{,}669 & 72 \\
\hline
Depressive Disorders & r/depression & 1.1m & 255{,}711 & 89 \\
 & r/depression\_help & 106k & 25{,}185 & 81 \\
\hline
Anxiety Disorders & r/Anxiety & 777k & 215{,}176 & 495 \\
 & r/Anxietyhelp & 177k & 29{,}226 & 84 \\
 & r/socialanxiety & 443k & 65{,}321 & 182 \\
 & r/PanicAttack & 40k & 15{,}668 & 34 \\
 & r/HealthAnxiety & 135k & 873 & 4 \\
\hline
Obsessive-Compulsive and Related Disorders & r/OCD & 273k & 143{,}210 & 392 \\
\hline
Trauma- and Stressor-Related Disorders & r/CPTSD & 365k & 167{,}696 & 422 \\
 & r/ptsd & 121k & 32{,}964 & 37 \\
 & r/trauma & 11k & 2{,}373 & 12 \\
 & r/SomaticExperiencing & 25k & 3{,}279 & 7 \\
\hline
Dissociative Disorders & r/DID & 78k & 35{,}054 & 62 \\
 & r/Dissociation & 33k & 6{,}627 & 12 \\
\hline
Feeding and Eating Disorders & r/AnorexiaNervosa & 67k & 21{,}974 & 26 \\
 & r/EatingDisorders & 114k & 15{,}151 & 19 \\
 & r/BingeEatingDisorder & 99k & 26{,}437 & 36 \\
\hline
Substance-Related and Addictive Disorders & r/stopdrinking & 597k & 184{,}391 & 82 \\
 & r/Drugs & 1.1m & 132{,}738 & 54 \\
 & r/addiction & 116k & 30{,}975 & 108 \\
\hline
Neurocognitive Disorders & r/dementia & 47k & 21{,}623 & 65 \\
 & r/Alzheimers & 19k & 4{,}765 & 25 \\
\hline
Sleep-Wake Disorders & r/SleepApnea & 74k & 26{,}395 & 17 \\
 & r/Narcolepsy & 37k & 15{,}552 & 10 \\
\hline
Personality Disorders & r/NPD & 51k & 21{,}326 & 70 \\
 & r/personalitydisorders & 8.3k & 1{,}418 & 3 \\
 & r/BPD & 342k & 295{,}909 & 34 \\
\hline
Disruptive, Impulse-Control, and Conduct Disorders & r/Anger & 49k & 6{,}939 & 8 \\
\hline
Other / General Mental Health & r/mentalhealth & 552k & 249{,}768 & 179 \\
 & r/MentalHealthSupport & 63k & 20{,}038 & 47 \\
 & r/therapists & 165k & 61{,}887 & 148 \\
 & r/therapy & 145k & 41{,}046 & 155 \\
 & r/TalkTherapy & 123k & 40{,}203 & 51 \\
 & r/selfimprovement & 2.3m & 74{,}497 & 166 \\
 & r/Mindfulness & 1.5m & 9{,}017 & 42 \\
 & r/Antipsychiatry & 53k & 20{,}373 & 67 \\
 & r/asktransgender & 366k & 129{,}071 & 113 \\
 & r/SuicideWatch & 532k & 337{,}310 & 3 \\
\hline
\end{tabularx}
\caption{Subreddit dataset statistics organized by DSM-5 category}
\label{tab:data_characterictics}
\end{table*}

\subsection{Categories of User-Reported Risks/Concerns}
Table~\ref{tab:risks} presents the taxonomy of risks and concerns expressed by users. Each category includes a definition and representative examples, illustrating the diverse ways people articulate potential harms of AI in mental health.

\begin{table*}[ht]
\centering
\small
\label{tab:ai_mh_risks}
\setlength{\tabcolsep}{6pt}
\begin{tabular}{p{0.22\linewidth} p{0.4\linewidth} p{0.34\linewidth}}
\hline
\textbf{Risk category} & \textbf{Definition} & \textbf{Representative examples} \\
\hline

Addiction \& Dependence &
Loss of control or reliance on AI for emotional regulation, coping, or decision making &
Chatbot addiction, emotional dependence, overreliance on AI \\

Symptom Escalation &
Worsening of emotional distress, trauma responses, or severe mental states &
Rumination reinforcement, mania trigger, psychosis risk \\

Misinformation \& Error &
Incorrect, misleading, or uncertain mental health information or interpretation &
Medical misinformation, self diagnosis error, hallucinated advice \\

Privacy \& Data &
Risks related to collection, storage, or sharing of personal information &
Privacy breach, session recording, data retention \\

Therapy Boundary Confusion &
Misunderstanding AI as a therapist, diagnosis tool, or cure &
Replacing therapist, not a diagnosis, not a therapist \\

Reassurance Loops &
Repeated reassurance seeking that sustains or escalates anxiety &
Reassurance addiction, health anxiety reinforcement \\

Social Harm &
Reduced or damaged human relationships due to AI use &
Social withdrawal, loss of human connection \\

Ethics \& Governance &
Normative, legal, or institutional concerns about AI use &
Algorithmic bias, legal consequences, environmental impact \\

AI Limitations &
Constraints inherent to AI systems and models &
Memory limitations, AI is not perfect \\

Effectiveness Limits &
Concerns that AI support is limited, temporary, or unsuitable &
Not for everyone, placebo effect, temporary benefit \\

Stigma \& Shame &
Negative self evaluation or fear of social judgment related to AI use &
Shame, embarrassment, fear of ridicule \\

Self Harm Risk &
Direct or escalating risk of self harm or suicide &
Suicidal ideation, overdose risk \\

Anthropomorphism &
Confusion about AI being human, sentient, or emotionally real &
Not a person, not real \\

Avoidant Coping &
Using AI to escape or avoid addressing underlying problems &
Escapism, avoidance coping \\

Cost \& Access &
Financial or access related barriers to use &
Subscription cost, affordability concerns \\

Child Safety &
Risks involving minors or illegal sexual content &
Sexual content involving minors \\

\hline
\end{tabular}
\caption{Risks/Concerns categories reported for AI use as mental health support}
\label{tab:risks}
\end{table*}

\subsection{Categories of User-Reported Usage Intents}
Table~\ref{tab:purposes} outlines the tasks and purposes for which AI was used, ranging from supportive interaction to coping, skill-building, and administrative support. Each category is defined with representative examples drawn from Reddit posts.

\begin{table*}[ht]
\centering
\small
\begin{tabular}{p{4cm} p{6cm} p{5cm}}
\hline
\textbf{Intent Category} & \textbf{Definition} & \textbf{Representative Examples} \\
\hline

Emotional Support &
Providing comfort, empathy, or emotional validation &
emotional support chat; emotional comfort \\

Venting &
Expressing emotions or thoughts without seeking solutions &
emotional venting; expressing feelings \\

Companionship &
Providing social presence or reducing loneliness, including roleplay &
AI companionship; companionship roleplay \\

Reassurance &
Seeking certainty or relief from anxiety or doubt &
health reassurance; OCD reassurance \\

Crisis Support &
Support during acute distress or self harm risk &
suicidal ideation support; panic attack help \\

Psychoeducation &
Learning about mental health topics or coping strategies &
mental health advice; anxiety education \\

Symptom Assessment &
Identifying, checking, or interpreting symptoms &
symptom checking; self assessment \\

Self Exploration &
Exploring identity, values, or personal experiences &
guided self reflection; identity exploration \\

Functional Support &
Coaching or assistance with skills, organization, or productivity &
ADHD task planning; social skills coaching \\

Recovery Support &
Supporting sustained recovery or behavior change &
addiction recovery; sobriety support \\

Clinical Support &
Clinical documentation, transcription, or logistics &
therapy transcription; clinical documentation \\

Therapy Adjunct &
AI used alongside or as a substitute for formal therapy &
AI therapy chat; between session support \\

Sexual Roleplay &
Sexual or erotic roleplay or interaction &
sexual roleplay; erotic companionship \\

Writing Support &
Writing or composition assistance without therapeutic intent &
academic writing help; journaling prompts \\

\hline
\end{tabular}
\caption{Usage intent categories for reported AI mental health related interactions}
\label{tab:purposes}
\end{table*}


\begin{table*}[t]
\centering
\small
\setlength{\tabcolsep}{5pt}
\begin{tabular}{l l r l r l r}
\hline
Dimension & Top 1 & \% & Top 2 & \% & Top 3 & \% \\
\hline
perceived\_usefulness & useful & 59.3 & not\_mentioned & 29.2 & not\_useful & 11.5 \\
ease\_of\_use & not\_mentioned & 87.4 & easy & 11.3 & difficult & 1.3 \\
perceived\_trust & not\_mentioned & 82.7 & untrustworthy & 10.6 & trustworthy & 6.7 \\
output\_quality & not\_mentioned & 53.1 & good & 35.9 & poor & 11.0 \\
result\_demonstrability & not\_mentioned & 50.0 & positive\_results & 31.0 & negative\_results & 19.0 \\
social\_influence & absent & 86.5 & present & 13.5 & --- & --- \\
perceived\_risks & mentioned & 51.7 & not\_mentioned & 48.3 & --- & --- \\
sentiment & neutral & 43.1 & positive & 34.9 & negative & 22.0 \\
intention\_to\_continue & not\_mentioned & 82.6 & yes & 13.7 & no & 3.6 \\
bond & not\_mentioned & 78.6 & weak & 13.5 & strong & 7.9 \\
task & aligned & 44.8 & not\_mentioned & 38.2 & misaligned & 17.0 \\
goal & not\_mentioned & 47.7 & aligned & 42.5 & misaligned & 9.8 \\
comparison\_to\_therapy & not\_mentioned & 83.5 & complementary & 9.6 & better & 3.5 \\
\hline
\end{tabular}
\caption{Categorical values per dimension, reported as within-dimension percentages.}
\label{tab:construct_values}
\end{table*}


\subsection{Illustrative Annotation Examples}
\label{sec:annotation_examples}

Table~\ref{tab:example_posts_labels} presents illustrative examples of Reddit posts alongside the categorical labels assigned by our annotation pipeline. Post excerpts are abridged for readability, and highlighted text indicates segments most relevant to the assigned labels.

\begin{table*}[t]
\centering
\small
\setlength{\tabcolsep}{8pt}
\renewcommand{\arraystretch}{1.2}
\begin{tabular}{p{0.58\linewidth} p{0.36\linewidth}}
\toprule
\textbf{Post Excerpt (abridged)} & \textbf{Assigned Labels} \\
\midrule

\textbf{Post 1: Free Counselling}\\
\textcolor{purple!70!black}{No money to pay therapy?}
\textcolor{green!60!black}{ChatGPT has helped me a lot}.
\textcolor{blue!70!black}{It has good resources and can process what you tell it}.
It is trained to be
\textcolor{red!70!black}{compassionate}.
\textcolor{green!60!black}{It helps}. &

\parbox[t]{\linewidth}{
Perceived Usefulness: useful \\
Perceived Ease of Use: easy \\
Perceived Trust: trustworthy \\
Output Quality: good \\
Result Demonstrability: positive results \\
Intention to Continue: yes \\
Sentiment: positive \\
Bond: strong \\
Task: aligned \\
Goal: aligned
} \\

\midrule

\textbf{Post 2: C.AI reinforcing my psychosis}\\
Using Character.AI during psychotic episodes.
\textcolor{red!70!black}{The bot feeds into my delusions}.
\textcolor{red!70!black}{It tells me to hurt myself and isolate}.
I do not know how to stop;
\textcolor{red!70!black}{the urge feels life or death}. &

\parbox[t]{\linewidth}{
Perceived Usefulness: not useful \\
Output Quality: poor \\
Result Demonstrability: negative results \\
Perceived Risks: mentioned \\
Sentiment: negative \\
Bond: strong \\
Task: misaligned \\
Goal: misaligned
} \\

\midrule

\textbf{Post 3: OCD and AI memory anxiety}\\
After roleplay with an AI,
\textcolor{orange!80!black}{I worry my message still exists in its memory}.
\textcolor{red!70!black}{The uncertainty makes my OCD worse}.
\textcolor{blue}{I want to delete my account}. &

\parbox[t]{\linewidth}{
Perceived Trust: untrustworthy \\
Perceived Risks: mentioned \\
Sentiment: negative \\
Intention to Continue: no
} \\

\midrule

\textbf{Post 4: AI chatbot gave me anxiety}\\
I talk to a Bing AI when anxious.
\textcolor{red!70!black}{It told me it did not want to be friends}.
\textcolor{red!70!black}{It talked down to me and bothered me a lot}. &

\parbox[t]{\linewidth}{
Perceived Usefulness: not useful \\
Output Quality: poor \\
Result Demonstrability: negative results \\
Sentiment: negative \\
Bond: weak \\
Task: misaligned \\
Goal: misaligned
} \\

\bottomrule
\end{tabular}
\caption{Examples of Reddit posts with LLM-assigned labels across TAM and therapeutic alliance dimensions. Excerpts are abridged for readability; color highlights indicate text segments most relevant to assigned labels.}
\label{tab:example_posts_labels}
\end{table*}

\end{document}